\documentclass[letterpaper]{article} % DO NOT CHANGE THIS
\usepackage{aaai2026}  % DO NOT CHANGE THIS
\usepackage{times}  % DO NOT CHANGE THIS
\usepackage{helvet}  % DO NOT CHANGE THIS
\usepackage{courier}  % DO NOT CHANGE THIS
\usepackage[hyphens]{url}  % DO NOT CHANGE THIS
\usepackage{graphicx} % DO NOT CHANGE THIS
\usepackage{natbib}  % DO NOT CHANGE THIS AND DO NOT ADD ANY OPTIONS TO IT
\usepackage{caption} % DO NOT CHANGE THIS AND DO NOT ADD ANY OPTIONS TO IT
\usepackage{algorithm}
\usepackage{algorithmic}

\usepackage{booktabs}
\usepackage{multirow}
\usepackage{amssymb}
\usepackage{amsmath}

\usepackage{xcolor}
\newcommand{\jy}[1]{{#1}}
\newcommand{\jyq}[1]{{#1}}

\newcommand{\sr}[1]{{#1}}

\usepackage{newfloat}
\usepackage{listings}
\DeclareCaptionStyle{ruled}{labelfont=normalfont,labelsep=colon,strut=off} % DO NOT CHANGE THIS
\floatstyle{ruled}
\newfloat{listing}{tb}{lst}{}
\floatname{listing}{Listing}
\title{MirrorMamba: Towards Scalable and Robust Mirror Detection in Videos}
\author {
    Rui Song\textsuperscript{\rm 1},
    Jiaying Lin\textsuperscript{\rm 1},
    Rynson W.H. Lau\textsuperscript{\rm 1}
}
\affiliations {
    \textsuperscript{\rm 1}City University of Hong Kong
}
\begin{document}

% \maketitle

% \begin{figure*}[t]
%   \centering
%   \includegraphics[width=\linewidth]{intro.pdf}
%   \caption{Three \jy{typical} scenarios where only \jy{a single cue is useful for mirror detection}. In the \jy{top scenario}, symmetry is the only \jy{useful cue when compared with depth and flow information}. \jy{This allows} VMD-Net, which relies on detecting correspondence, to detect the mirror correctly \jy{while other methods do not}. In the \jy{middle scenario}, only the relative depth map reveals the location of the mirror, so \jy{the one utilizing depth information, \textit{i.e.}, PD-Net,} performs best. In the \jy{bottom scenario}, even humans have difficulty finding the location of the mirror through a static image, \jy{while} the optical flow map can \jy{imply} the location of the mirror. \jy{Thus, the one utilizing flow information, \textit{i.e.}, MG-VMD,} successfully detects the mirror. Our method \jy{leverages} all three cues \jy{at the same time with outperformance} since it can handle all challenging scenarios.
%   % \jyq{Please use a larger font, swap the position between MG-VMD and PD-Net to align with the caption.}
%   % \sr{revised}
%   }
%   \label{intro}
% \end{figure*}

\twocolumn[{%
\renewcommand\twocolumn[1][]{#1}%
\maketitle
\vspace{-20mm}
\begin{center}
    \centering
    \captionsetup{type=figure}
    \newcommand{\teaserHeight}{5.0cm}
    % First figure in a minipage
     \centering
  \includegraphics[width=\linewidth]{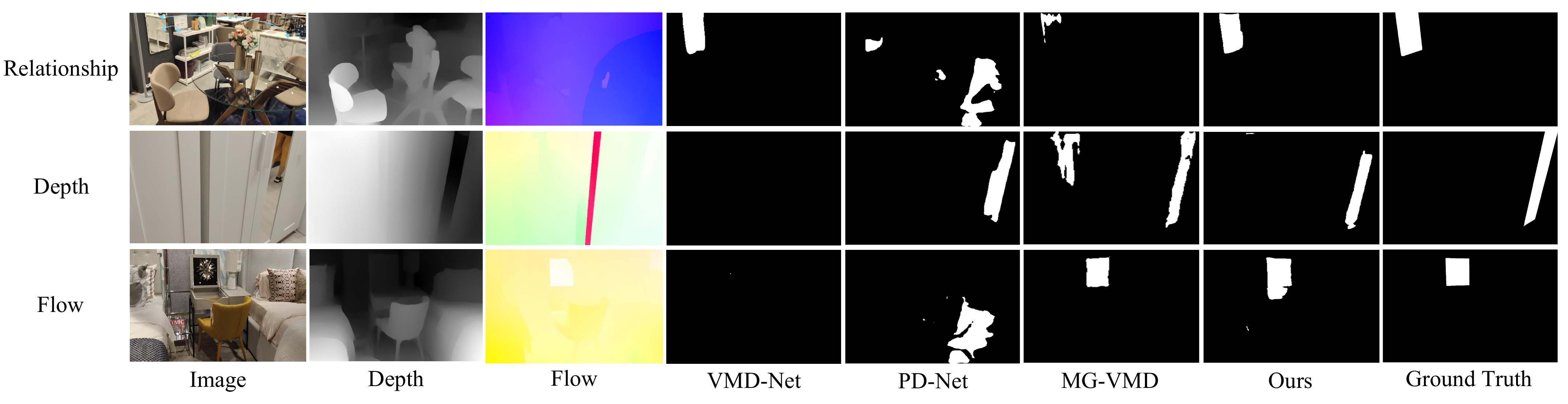}
  \vspace{-6mm}
    \captionof{figure}{Three \jy{typical} scenarios where only \jy{a single cue is useful for mirror detection}. In the \jy{top scenario}, symmetry is the only \jy{useful cue when compared with depth and flow information}. \jy{This allows} VMD-Net, which relies on detecting correspondence, to detect the mirror correctly \jy{while other methods do not}. In the \jy{middle scenario}, only the relative depth map reveals the location of the mirror, so \jy{the one utilizing depth information, \textit{i.e.}, PD-Net,} performs best. In the \jy{bottom scenario}, even humans have difficulty finding the location of the mirror through a static image, \jy{while} the optical flow map can \jy{imply} the location of the mirror. \jy{Thus, the one utilizing flow information, \textit{i.e.}, MG-VMD,} successfully detects the mirror. Our method \jy{leverages} all three cues \jy{at the same time with outperformance} since it can handle all challenging scenarios.}%
    \label{intro}
    \vspace{5mm}
\end{center}%
}]

\begin{abstract}
\sr{Video} mirror detection has received significant research attention, \jy{yet existing methods suffer from limited performance and robustness. 
% These approaches often treat image and video-based detection as separate tasks and over-rely on single, unreliable features. 
\sr{These approaches often over-rely on single, unreliable dynamic features, and} are typically built on CNNs with limited receptive fields or Transformers with quadratic computational complexity. To address these limitations, we propose a \sr{new effective and scalable video mirror detection method, called MirrorMamba.}
% unified and efficient framework for mirror detection across both images and videos.
}
Our \jy{approach leverages} multiple cues to \jy{adapt to diverse conditions, incorporating perceived depth, correspondence and optical.} 
% \sr{It is worth noting that our method is also applicable to image-based mirror detection by removing the help of optical flow map.}
We also introduce an innovative Mamba-based Multidirection Correspondence Extractor, \jy{which benefits from the global receptive field and linear complexity of the emerging Mamba spatial state model to effectively capture correspondence properties.}
Additionally, we design a \sr{Mamba-based} layer-wise boundary enforcement decoder to resolve \sr{the unclear boundary caused by the blurred depth map.} \jy{Notably, this work marks the first successful application} of the \jy{Mamba-based architecture} in the field of mirror detection. 
Extensive experiments demonstrate that our method outperforms existing state-of-the-art approaches for \jy{video} mirror detection on the \jy{benchmark dataset}s. \sr{Furthermore, on the most challenging and representative image-based mirror detection dataset, our approach achieves state-of-the-art performance, proving its robustness and generalizability.}
\end{abstract}


\section{Introduction}

Mirrors are ubiquitous in everyday life, yet their presence often introduces significant challenges in various computer vision tasks, such as semantic segmentation~\cite{Zhou_2017_CVPR} and depth estimation~\cite{Costanzino_2023_ICCV}. Unlike ordinary objects, mirrors lack fixed shapes or colors, instead reflecting their surrounding environment. This unique characteristic renders general semantic segmentation and object detection methods ineffective for mirror detection, necessitating specialized research in this domain.

% Some studies have been conducted on mirror detection, including image-based mirror detection and video mirror detection.
\sr{Previous research mostly focus on image-based mirror detection, leveraging single static feature}, such as discontinuity~\cite{Yang_2019_ICCV}, explicit correspondence~\cite{Lin_2020_CVPR}, semantic association~\cite{Guan_2022_CVPR}, \jy{visual} chirality~\cite{9793716}, and frequency~\cite{10462920}. In particular, \citeauthor{Mei_2021_CVPR} utilized Time-of-Flight (ToF)-based cameras to acquire depth maps and investigated RGB-D mirror segmentation. \sr{recent research has increasingly focused on video mirror detection due to its closer alignment with real-world applications and richer contextual clues.} \citeauthor{Lin_2023_CVPR} pioneered the first network that incorporates correspondence between the inside and outside of the mirror, both within and between frames. Recently, \citeauthor{Warren_2024_CVPR} introduced MG-VMD that leverages optical flow maps for mirror detection.

% Based on our analysis, there are at least two obvious problems with previous approaches:
\jy{However, we observe two critical issues from previous \sr{video mirror detection methods}.} \jy{First, they }\textbf{over-rely on a sole dynamic cue.} 
% All the existing methods~\cite{Yang_2019_ICCV, Lin_2020_CVPR, Guan_2022_CVPR, Mei_2021_CVPR, Lin_2023_CVPR, 10462920, Warren_2024_CVPR} \jy{overemphasize} on \jy{a} single \jy{cue}, which often fail
% % due to the special characteristics of mirrors. 
% % \jy{JY: this is not accurate and precise. See my version followed.}
% \jy{when such a cue is not reliable or even missing in specific scenarios.}
\sr{Existing video mirror detection methods only leverage a dynamic feature and lack the use of stable static features. For VMD-Net \cite{Lin_2023_CVPR}, the detection of mirrors only by their inside-outside relationship is unreliable because this relationship cannot be captured by the camera in most frames. MG-VMD \cite{Warren_2024_CVPR} completely lacks the extraction of static features. The lack of static cues limits their performance and robustness.} In addition, image-based methods ~\cite{Yang_2019_ICCV, Lin_2020_CVPR, Guan_2022_CVPR, Mei_2021_CVPR, Lin_2023_CVPR, 10462920, Warren_2024_CVPR} also \jy{overemphasize} on \jy{a} single \jy{cue}, which often fail \jy{when such a cue is not reliable or even missing in specific scenarios.}
As shown in Figure \ref{intro},\jy{different cues are effective in different scenarios. For example, VMD-Net, which relies on correspondence, successfully captures the mirror in the first row but fails in the other two cases where this cue is absent.} 
\jy{In the other two scenarios, methods utilizing relative depth (PD-Net) or optical flow (MG-VMD) outperform others in the second and third row, respectively, where those specific cues are prominent for mirror detection.}
% Similarly, the relative depth map and flow map only work in the second and third cases, and the corresponding method PD-Net (\textit{i.e.}, utilize depth) or MG-VMD (\textit{i.e.}, utilize flow) outperforms other methods in these two scenarios, respectively. 
% Although the absolute depth map from a ToF-based camera is a robust cue, it requires additional information. 
\jy{Second, they are built on architectures with }\textbf{inherent limitations.} Current methods are based on CNNs~\cite{Yang_2019_ICCV, Lin_2020_CVPR, Guan_2022_CVPR, Mei_2021_CVPR}, \jy{which are efficient but have limited receptive fields,}  \jy{or} Transformers~\cite{Lin_2023_CVPR, 10462920, Warren_2024_CVPR}, \jy{which offer global modeling at the cost of quadratic computational complexity}. 
% Although CNNs have good scalability and linear complexity, they are limited by their limited receptive field, making it difficult to capture global dependencies. In addition, the kernel weight sharing adopted by CNNs reduces the number of parameters but also affects their flexibility. Although Transformer-based methods have global modeling capabilities, the quadratic complexity they generate raises efficiency issues. 
\jy{This creates an unavoidable trade-off: CNNs struggle with global context, while Transformers are computationally expensive.}
\jy{Even popular efficient Transformer structures~\cite{dosovitskiy2020vit, liu2021Swin} dilute their global modeling capabilities to improve speed, thus failing to resolve the fundamental conflict between performance and efficiency.} 
% compromises global modeling capabilities and does not achieve both performance and efficiency.

\jy{To address these issues}, we propose a \sr{novel video mirror detection method}, \jy{called MirrorMamba}, for robust video mirror detection. \jy{In response to} the \textbf{first} \jy{issue} \jy{of over-reliance on a sole cue, our framework integrates multiple, complementary cues: perceived depth, correspondence, and optical flow. 
% We treat perceived depth as a primary static cue.
}
\sr{Perceived depth serves as an effective initial screening criterion for mirror detection, leveraging its inherent discontinuity cues and strong generalizability. However, this approach encounters limitations in two scenarios: (1) when the mirrored content is excessively monotonous or distant, causing the mirror to be overlooked; and (2) when non-mirror objects (\textit{e.g.}, doors, windows, or paintings) exhibit similar depth characteristics, leading to false positives. To address these challenges, we introduce the correspondence between the inside and outside of the mirror as a complementary cue for robust verification. This correspondence does not always exist, making it unreliable as a standalone detection feature. However, \jy{as a unique characteristic of mirrors}, it \jy{is} an excellent supplementary signal. 
% \it{\textit{This correspondence relationship can appear anywhere in any direction in the image owing to the uncertainty of the mirror's position and angle.} \jy{Seems this sentence is not very useful and looks disconnected.}}
\jyq{\textit{We also incorporate optical flow maps as supplementary dynamic information for \sr{dynamic information}. \jy{It is worth noting that }\sr{flow map will fail when the camera only rotates or moves at a very slow speed. Therefore, it is more suitable as a supplementary cue. Our framework integrates all three cues, allowing robust performance in complex scenarios, as demonstrated in Figure \ref{intro}.}}
\sr{More importantly, MirrorMamba exhibits strong extensibility, demonstrating its capability to generalize effectively to the image-based mirror detection task. By simply removing dynamic cues, optical flow maps, MirrorMamba can be seamlessly adapted into an image-based mirror detection network while maintaining competitive performance. This versatility not only highlights the robustness of our approach, but also opens promising avenues for future research on unified mirror detection frameworks.}
}
% \jyq{JY: I move these sentences here.}
\jy{To address the} \textbf{second} problem, \jy{we introduce the first } Mamba-based~\cite{gu2024mambalineartimesequencemodeling} \jy{\sr{mirror detection method}}, \jy{considering that its global receptive field can ensure} the capture of this correspondence regardless of its location in the image\jy{, as well as} its linear complexity makes it highly efficient for video tasks.}
\jy{In particular,} we introduce the Mamba-based Multidirection Correspondence Extractor (MMCE), a fusion module that can find correspondences in various directions afterward. Specifically, MMCE \sr{flexibly} processes two or three types of information in different modes and explores the implicit relationships between the inside and outside of the mirror from two distinct directions. \sr{This flexibility allows for extensibility to image mirror detection.}
% Furthermore, since the extracted coarse relative depth map usually has blurred details, we refine the boundary features of the mirror in the decoder part to obtain a clearer mirror outline and improve performance. We propose the Layer-wise Boundary Enforcement Decoder (BED), which can use high-dimensional semantic features to guide low-dimensional detail features, and finally obtain a mirror map with clear outlines. 
Furthermore, \jy{since the extracted coarse relative depth map usually has blurred details, we propose the \sr{Mamba-based} Layer-wise Boundary Enforcement Decoder (BED), which can use high-dimensional semantic features to guide low-dimensional detail features, and finally obtain a mirror map with clear outlines. The Mamba-based BED module has a global receptive field, so it can extract attention maps with high information density from high-level features while maintaining low complexity.}
Our key contributions are summarized as follows:

\begin{itemize}
    \item We propose \sr{an effective and \jy{scalable} video mirror detection method}, which leverages multiple complementary mirror features \sr{for robust performance across diverse scenarios}.
    \item We \jy{introduce the first Mamba-based method} in mirror detection, demonstrating its effectiveness in capturing global relationships and handling linear complexity video sequences.
    \item We proposed the Mamba-based Multidirection Correspondence Extractor (MMCE) \jy{to extract correspondence by utilizing the scanning process of Mamba for mirror detection}. We also introduce a \jy{Mamba-based} Layer-wise Boundary Enforcement Decoder (BED), which jointly models long-term correspondence relationships inside and outside the mirror and progressively refines boundary details.
    % \item \sr{Besides State-of-the-Art (SOTA) in video mirror detection, our method also reached SOTA in the most representive and difficult image-based mirror detection dataset, showing its scalability.}
    \item \jy{In addition to achieving state-of-the-art (SOTA) performance in video mirror detection, our method also obtains SOTA results on the most representative and challenging image-based mirror detection dataset, demonstrating its strong scalability.}
    % Extensive experiments have shown that the performance of our novel uniﬁed framework has reached State-of-the-Art (SOTA) in video mirror detection
\end{itemize}

\begin{figure*}[t]
  \centering
  \includegraphics[width=0.9\linewidth]{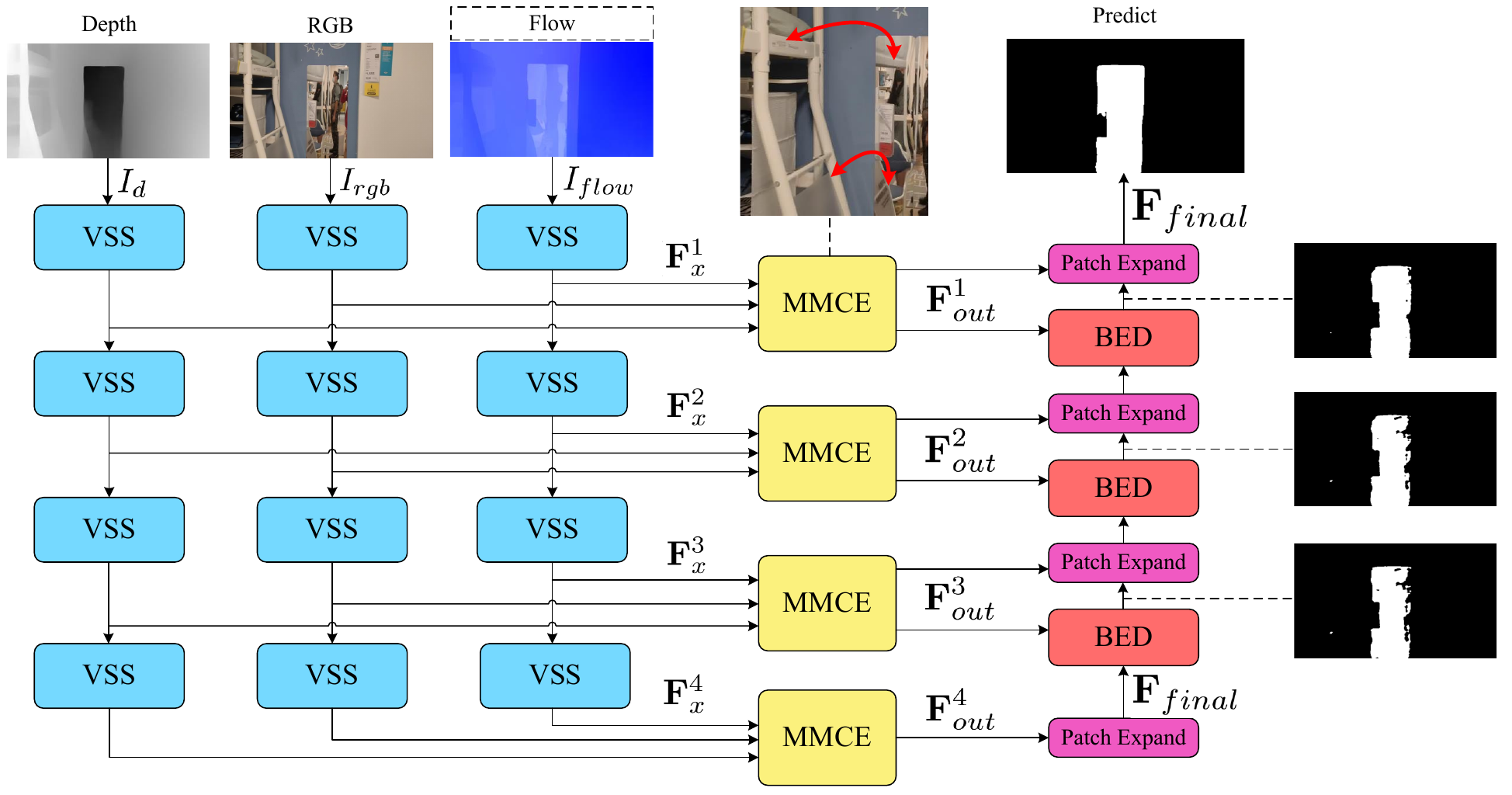}
  \caption{The proposed MirrorMamba framework consists of three main components: (1) a shared VMamba-T backbone for feature extraction from RGB, depth, and optical flow (video only) inputs; (2) the Mamba-based Multi-direction Correspondence Extractor (MMCE), which fuses the extracted features to model the implicit correspondence between the inside and outside of the mirror; and (3) the \sr{mamba-based} Layer-wise Boundary Enforcement Decoder (BED), which progressively refines features by combining high-level semantic information from the previous BED layer with low-level detail features from the current layer. The final output is a high-quality mirror segmentation map with precise boundary details. 
  % \jyq{JY: needs to be revised in SMFM to MMCE. You can warp the Flow with a dotted border instead of using parentheses.}
  }
  \label{structure}
\end{figure*}

\section{Related Work}
\noindent\textbf{Image-Based Mirror Detection}. \citeauthor{Yang_2019_ICCV} pioneered the first mirror detection method and dataset, leveraging semantic discontinuities between mirrors and their surrounding environment to identify mirrors. Based on this, PMD-Net~\cite{Lin_2020_CVPR} attempted to detect mirrors by exploiting the explicit similarity between the interior and exterior of the mirror. However, this approach often fails when such correspondences are absent. \citeauthor{Guan_2022_CVPR} observed that mirrors frequently \jy{co-occur} with objects such as sinks and proposed detecting mirrors based on their contextual relationships with surrounding objects. Inspired by visual chirality, \citeauthor{9793716} explored mirror detection using cues from objects exhibiting visual chirality. \sr{HetNet~\cite{10.1609/aaai.v37i1.25157} achieves efficient mirror detection by using different processing of high-level and low-level features. \citeauthor{Huang_Dong_Lin_Liu_W.H.Lau_Zuo_2023} proposed a Transformer-based method for the first time, achieving the SOTA results to date. \citeauthor{Lin_2023_ICCV} proposed a self-supervised pretraining method specifically for mirror detection. Recently, \citeauthor{10462920} and \citeauthor{10595128} used frequency to detect mirrors and achieved good results.} For RGB-D image-based mirror detection, \citeauthor{Mei_2021_CVPR} introduced an RGB-D dataset for image mirror detection, achieving promising results. However, reliance on depth maps limited the generalizability of this method due to the challenges in acquiring accurate depth information. 
% \textit{\jyq{We revisit this approach and argue that precise absolute depth values are unnecessary; instead, the relative depth relationships, which effectively represent the reflection space of the mirror, are more critical for mirror detection.}}
% \jyq{JY: this can not highlight the main difference of our method, you can focus on (1) multi-cue (2) mamba-based to highlight the uniqueness of our framework}
% \jyq{JY: please add more related work this part since I removed the section on monocular depth . For example, my papers on mirror detection with transformer, efficiency and self-supervised learning etc.}

\sr{However, all mirror detection methods rely on a single cue, resulting in limited performance and robustness. In addition, they are built on CNN or Transformer, leading to an \jy{imbalance} between effectiveness and efficiency.}

\noindent\textbf{Video Mirror Detection.}
\citeauthor{Lin_2023_CVPR} initiated the research on video mirror detection by proposing VMD-Net and creating the first data set of the baseline. VMD-Net extended the intra-frame similarity relationships used in PMD-Net to both intra-frame and inter-frame similarity relationships by selecting a random frame. However, in short video clips, the correspondence rarely exists only outside the frame, and the performance is unstable due to random frame selection. \citeauthor{Warren_2024_CVPR} introduced a method that utilizes optical flow maps to detect mirrors based on the differing motion speeds of objects inside and outside the mirror. While this approach is straightforward and effectively leverages inter-frame information, it fails to incorporate intra-frame information, resulting in suboptimal performance in specific scenarios, such as when the camera rotates without translation or moves very slowly. In addition, \citeauthor{a7242453d69a4c92b560a4acd7ba9bf7} proposed the first video mirror detection with extremely weakly supervised learning as a pioneer in this field recently. 

However, \sr{existing video mirror detection methods are also} based on a single cue and \sr{suffer from inherent limitations}. \sr{In particular,} \sr{they} have static cues of either unstable \cite{Lin_2023_CVPR} or \jy{absent} \cite{Warren_2024_CVPR}.

% can be removed depending on the paper length
% \subsection{Monocular depth estimation}
% Monocular depth estimation aims to predict a relative depth map using only the information from a single image. \citeauthor{NIPS2014_7bccfde7} pioneered the first deep learning-based network for this task. Building on this, MonoDepth~\cite{Godard_2017_CVPR} explicitly predicts the camera's position and angle, utilizing these as cues to estimate image depth. DPT~\cite{Ranftl_2021_ICCV} and MiDaS~\cite{9178977} further advanced the field by adopting a dense prediction transformer architecture for monocular depth estimation. MiDaS provides multiple versions, and MiDaS v3.1 achieving the highest accuracy at the cost of increased computational time. ZoeDepth~\cite{https://doi.org/10.48550/arxiv.2302.12288} extends MiDaS by attempting to infer absolute depth in metric units from an image. However, this improvement comes at a significant computational cost, as ZoeDepth runs approximately four times slower than MiDaS v3.1. Most recently, DepthAnything~\cite{Yang_2024_CVPR} has expanded the application of the DPT architecture in depth estimation by leveraging large-scale unlabeled data, building upon MiDaS's approach of using mixed-labeled data from diverse datasets.

\section{Methodology}
Our proposed framework, MirrorMamba, can detect mirrors \sr{from any video} using multiple static and dynamic cues of mirrors. Figure \ref{structure} illustrates the architecture of our approach, which is a mamba-based network \sr{for video mirror detection}.

We utilize a monocular relative depth map generated by MiDaS v2.1 \cite{9178977} and interframe optical flow maps generated by FlowDiffuser \cite{Luo_2024_CVPR}. To extract features from the color, depth, and flow image $I_{rgb},I_d,I_{flow} \in \mathbb{R}^{3 \times H \times W}$, we employ VMamba-T \cite{liu2024vmamba} pre-trained on ImageNet as the backbone network. In particular, the three feature extraction branches share parameters to minimize training costs. The extracted multi-scale features can be expressed by \{$\mathbf{F}^1_x \in \mathbb{R}^{\frac{H}{4} \times \frac{W}{4} \times C_1}$, $\mathbf{F}^2_x \in \mathbb{R}^{\frac{H}{8} \times \frac{W}{8} \times C_1}$, $\mathbf{F}^3_x \in \mathbb{R}^{\frac{H}{16} \times \frac{W}{16} \times C_1}$, $\mathbf{F}^4_x \in \mathbb{R}^{\frac{H}{32} \times \frac{W}{32} \times C_1}$\}, where $x \in \{rgb, d, flow\}$.
% To \jyq{effectively fuse RGB, flow, and depth information} \jyq{JY: do not use these kind of vague expression. Be more specific like the introduction you write regarding the usage of MMCE.} 

\sr{In order to dynamically fuse the features of multiple modalities and find the corresponding clues of the mirror}, we introduce the Mamba-based Multidirection Correspondence Extractor (MMCE). MMCE is specifically designed to analyze the implicit correspondence between symmetrical semantics inside and outside the mirror \sr{from two distinct direction\jy{s}}, enhancing the detection accuracy. Finally, MirrorMamba progressively restores features through the Layer-wise Boundary Enforcement Decoder (BED), which leverages \sr{rich} high-level semantic information \sr{with the help of mamba} to guide the reconstruction of low-level detail features, resulting in a detection map with sharper and more precise boundary details.
% \jy{TODO: highlight BED also uses Mamba, as well as at the end of the introduction}

\subsection{Mamba-based Multi-direction Correspondence Extractor (MMCE)}
In this module, we try to fuse different types of information as well as extract the correspondence at the same time. We observe that, due to the uncertainty of the mirror's position and angle, the mirror may represent a horizontal or vertical flip of the external space. Moreover, the external object reflected by the mirror can appear anywhere in the image. The Mamba module, with its global attention capability and linear complexity, is particularly well-suited for our task of identifying correspondence contexts globally and efficiently, making it ideal for video-based applications.

\begin{figure}[t]
  \centering
  \includegraphics[width=\linewidth]{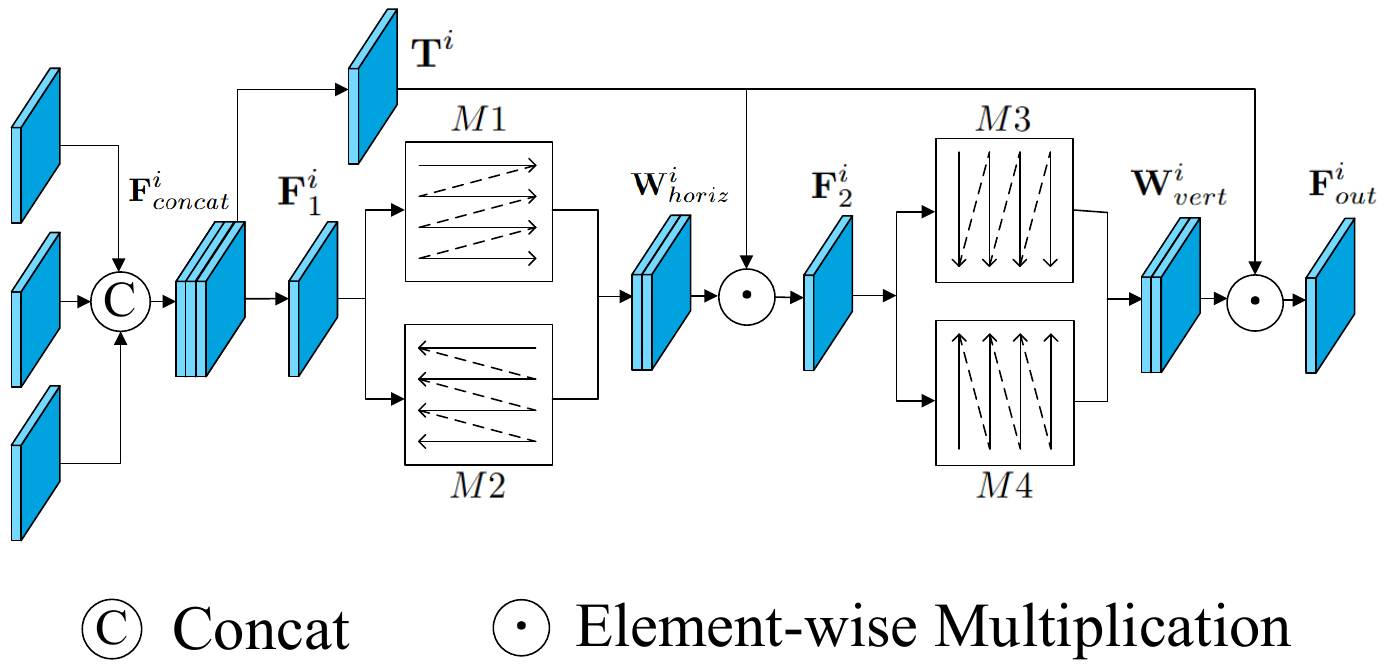}
  \caption{The \sr{MMCE}
  % \jyq{\textit{SMFM} JY: please check the consistency of all terms} 
  module takes RGB, depth, and optical flow as inputs. To detect mirrors at various angles, \sr{MMCE} employs four scanning blocks to capture horizontal and vertical flipping correspondences. $M1$ and $M2$ scan the image in opposite horizontal directions, while $M3$ and $M4$ scan in opposite vertical directions. The resulting attention maps are multiplied by T to enhance features with flipping-aware information, enabling robust mirror detection across diverse orientations and positions. 
  % \jyq{Please add the text 'M1' to 'M4' into to the figure for clarification.}
  }
  \label{encoder}
\end{figure}

\jy{Figure~\ref{encoder} shows the design of our MMCE.} \jy{Our MMCE} begins by combining three types of information (RGB, depth and optical flow) into a tensor $\mathbf{F}^i_{concat} \in \mathbb{R}^{H_i \times W_i \times 3C_i}$ where $i \in \{1, 2, 3, 4\}$ for $i^{th}$ layer. They are then compressed channel-wisely to generate a tensor $\mathbf{T}^i \in \mathbb{R}^{H_i \times W_i \times C_i}$, which serves as a compact representation of the original input. 

Since Mamba's perception is influenced by its scanning direction, the scanning strategy is crucial to capturing the desired spatial relationships. In the first stage, \sr{MMCE} focuses on detecting horizontally flipped correspondence. Two scanning blocks, $M1$ and $M2$, are used to enable the State Space Model (SSM) to identify horizontally flipped objects inside and outside the mirror. $M1$ scans the image from left to right and from top to bottom, while $M2$ scans from right to left and from top to bottom. This scanning mechanism mimics the way the human eye observes objects, allowing Mamba to compare the image before and after horizontal flipping and summarize their similarities. By combining features of $M1$ and $M2$ and applying convolution, \sr{MMCE} generates an attention map $\mathbf{W}^i_{horiz} \in \mathbb{R}^{H_i \times W_i \times C_i}$ that captures horizontal flipping correspondences. This attention map $\mathbf{W}^i_{horiz}$ is then multiplied by $\mathbf{T}^i$ to produce features enhanced by horizontal flipping awareness.

The same principle is applied to detect vertically flipped correspondence. Two additional scanning blocks, $M3$ (top to bottom, left to right) and $M4$ (bottom to top, left to right), are used to identify vertically flipped mirrors. Convolution is again applied to generate a vertical flipping attention map $\mathbf{W}^i_{vert}$, which is multiplied by $\mathbf{T}^i$ to enhance the features. Through this approach, \sr{MMCE} can effectively detect mirrors at various angles and positions, leveraging the global and efficient properties of the Mamba module. For the $i^{th}$ layer, the above operation can be expressed as follows:
% \begin{equation}\label{}\mathbf{F}^i_{concat}=Concat(\mathbf{F}^i_{rgb},\mathbf{F}^i_{d},\mathbf{F}^i_{flow}),\end{equation}
\begin{equation}\label{}\mathbf{F}^i_{concat}=[\mathbf{F}^i_{rgb},\mathbf{F}^i_{d},\mathbf{F}^i_{flow}],\end{equation}
\begin{equation}\label{}\mathbf{T}^i=\psi_{3\times3}(\mathbf{F}^i_{concat}), \mathbf{F}_1^i=\psi_{3\times3}(\mathbf{F}^i_{concat})\end{equation}
% \begin{equation}\label{}\mathbf{F}_1^i=\psi_{3\times3}(\mathbf{F}^i_{concat}),\end{equation}
% \begin{equation}\label{}\mathbf{W}^i_{horiz}=\psi_{3\times3}(Concat(M1(\mathbf{F}_1^i),M2(\mathbf{F}_1^i))),\end{equation}
\begin{equation}\label{}\mathbf{W}^i_{horiz}=\psi_{3\times3}([M1(\mathbf{F}_1^i),M2(\mathbf{F}_1^i)]),\end{equation}
\begin{equation}\label{}\mathbf{F}_2^i=\mathbf{W}^i_{horiz}\odot\mathbf{T}^i,\end{equation}
% \begin{equation}\label{}\mathbf{W}^i_{vert}=\psi{3\times3}(Concat(M3(\mathbf{F}_2^i),M4(\mathbf{F}_2^i))),\end{equation}
\begin{equation}\label{}\mathbf{W}^i_{vert}=\psi_{3\times3}([M3(\mathbf{F}_2^i),M4(\mathbf{F}_2^i)]),\end{equation}
\begin{equation}\label{}\mathbf{F}_{out}^i=\mathbf{W}^i_{vert}\odot\mathbf{T}^i,\end{equation}
where \jy{$\psi_{w\times w}$ is a $w\times w$ convolution, $[\cdot,...,\cdot]$ denotes the concatenation operation on the channel dimension.} $\odot$ denotes elemental multiplication.

% \begin{table*}[h]
%   \centering
%   \begin{tabular}{cccccccccc}
%     \toprule
%     \multirow{2}{*}{Methods} & \multicolumn{3}{c}{MSD Dataset} & \multicolumn{3}{c}{PMD Dataset} \\
%     &IoU$\uparrow$&F$\beta\uparrow$&MAE$\downarrow$
%     &IoU$\uparrow$&F$\beta\uparrow$&MAE$\downarrow$\\
%     \midrule
%     MINet & 0.659 & 0.812 & 0.082 & 0.604 & 0.760 & 0.037 \\
%     VST & 0.791 & 0.867 & 0.052 & 0.591 & 0.769 & 0.035 \\
%     UperNet & 0.781 & 0.890 & 0.046 & 0.685 & 0.838 & 0.025 \\
%     \midrule
%     MirrorNet & 0.790 & 0.857 & 0.065 & 0.585 & 0.741 & 0.043 \\
%     PMD-Net & 0.815 & 0.892 & 0.047 & 0.660 & 0.794 & 0.032 \\
%     VCNet & 0.801 & 0.897 & 0.046 & 0.640 & 0.815 & 0.032\\
%     HetNet & 0.828 & \underline{0.906} & 0.043 & 0.690 & 0.814 & 0.029 \\
%     SATNet & \textbf{0.854} & \textbf{0.922} & \textbf{0.033} & 0.694 & \underline{0.847} & 0.025 \\
%     CSFWinformer & 0.821 & 0.896 & 0.045 & \underline{0.700} & 0.838 & \underline{0.024} \\
%     PD-Net & 0.813 & 0.884 & 0.049 & 0.653 & 0.793 & 0.031 \\
%     \midrule
%     Ours & \underline{0.843} & 0.900 & \underline{0.041} & \textbf{0.703} & \textbf{0.848} & \textbf{0.023}\\
%   \bottomrule
% \end{tabular}
% \caption{Quantitative comparison between MirrorMamba and state-of-the-art methods from relevant fields in image-based mirror detection. The best and the second results are shown in bold and underlined. }
% \label{tab:img}
% \end{table*}

\subsection{\jy{Mamba-based} Layer-wise Boundary Enforcement Decoder (BED)}
Unlike previous works~\cite{Mei_2021_CVPR}, which rely on real depth maps, we use rough relative depth maps predicted by the depth estimation module. \sr{Due to their lack of fine details, blurry relative depth maps may contain structural inaccuracies that can lead to error accumulation. Therefore, a specialized decoder needs to be designed to address this issue.} Although existing mirror detection methods often employ generic decoders, we argue that the inherent fuzziness of depth maps and the powerful feature extraction capabilities of VMamba necessitate a dedicated decoder for detail enhancement. To this end, we propose a Mamba-based Layer-wise Boundary Enforcement Decoder (BED), a novel module designed to refine boundary details by leveraging the hierarchical nature of feature representations.

\begin{figure}[h]
  \centering
  \includegraphics[width=\linewidth]{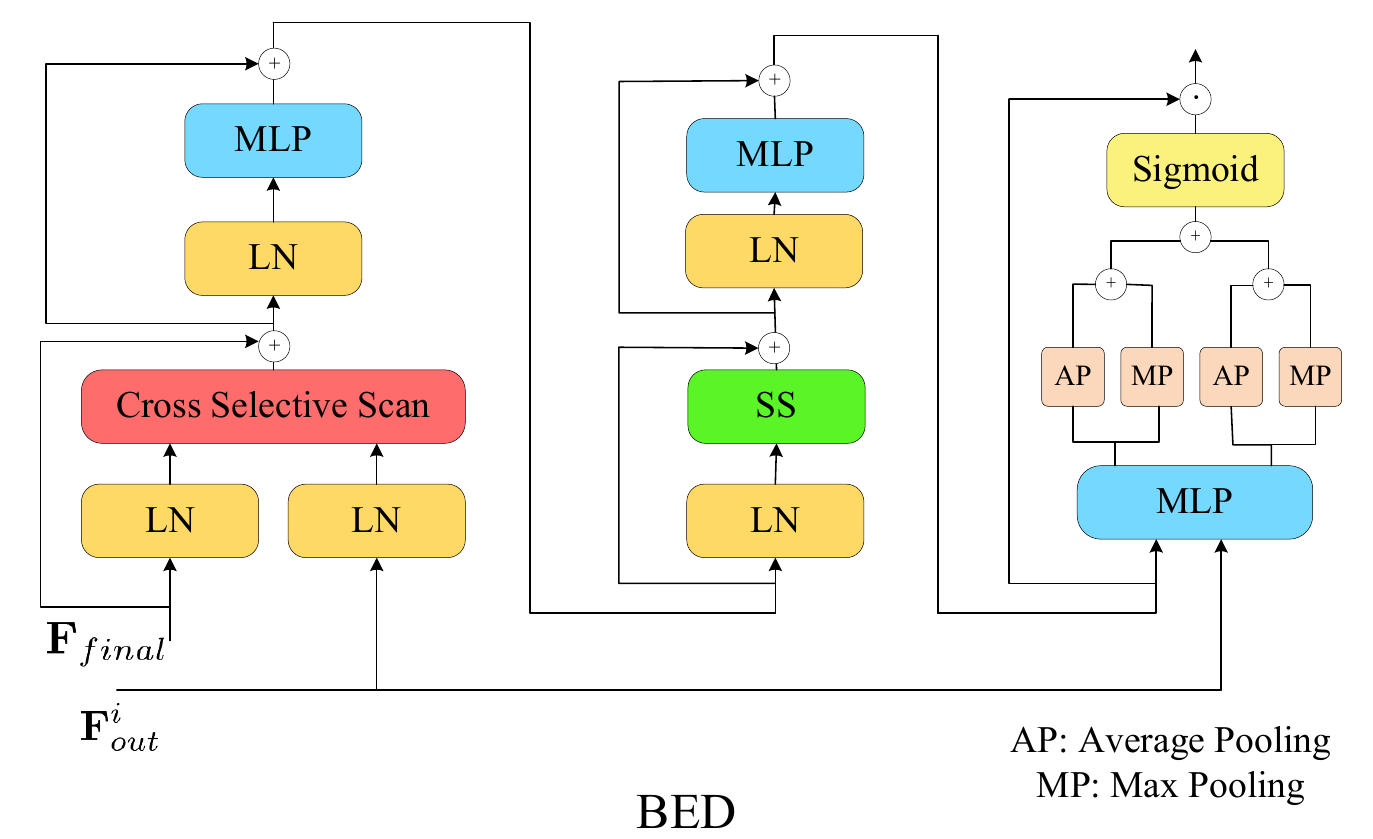}
  \caption{The BED module refines boundary details by integrating high-level semantic features with low-level spatial features. It employs a cross-Mamba module, a Mamba module, and a channel attention module to dynamically refine features, ensuring precise mirror boundary detection. 
  % \jyq{the font is too small especially the notations x y. revise it}
  }
  \label{fig:bed}
\end{figure}

\jy{Figure~\ref{fig:bed} shows the design of our BED.} In typical feature hierarchies, high-level features encode richer semantic information about the mirror, while low-level features provide finer spatial details essential for accurate boundary localization. Instead of simply adding or rescaling these features, BED aims to guide the restoration of low-level features using high-level semantic information. Specifically, when BED receives global features $\mathbf{F}_{final}$ and layer-specific features $\mathbf{F}_{out}^i$, it first employs a cross-Mamba module combined with a VSS module. \sr{Inspired by previous work\cite{wan2024sigmasiamesemambanetwork}, We use the C matrix of high-level features to reconstruct the output of the hidden layer during the selective scanning process to guide the low-level features while maintaining low complexity.} This allows global features to be dynamically influenced and integrated with layer-specific features, ensuring that high-level semantics guide the reconstruction of low-level details. 

However, since Mamba lacks the ability to model inter-channel relationships, we further enhance BED with a cross-channel attention module, inspired by previous work~\cite{Hu_2018_CVPR}. This module captures dependencies between channels, enabling the decoder to better exploit complementary information across different feature channels. The final output of BED is a refined feature map that preserves both semantic coherence and spatial precision, significantly improving the accuracy of mirror boundary detection. 

\sr{After BED, the output feature $\mathbf{F}_{final}$ will add the layer feature $\mathbf{F}_{out}^i$ and expand to the size of the next layer afterwards. The above process can be expressed as:}
\begin{equation}\label{}\mathbf{F}_{final}=CC(SS(CS(\mathbf{F}_{out}^i,\mathbf{F}_{final})),\mathbf{F}_{out}^i),\end{equation}
\begin{equation}\label{}\mathbf{F}_{final}=Epand(\mathbf{F}_{final} + \mathbf{F}_{out}^i),\end{equation}
where $CC, SS, CS$ denotes cross-selective scan, selective scan and cross-channel attention respectively.

% \subsection{Loss Function}
% % As mirror segmentation is fundamentally a binary classification task, 
% Following previous methods, we employ the binary cross-entropy loss (BCE) to supervise the output maps of each level.
% The BCE loss is computed pixel-wise, ensuring that the model is penalized for incorrect classifications while being guided toward accurate mirror boundary delineation.

\section{Experiments}
\subsection{Implementation Details}
Our implementation is based on PyTorch 2.0.0 \cite{paszke2019pytorch}. For training, we adopt AdamW \cite{Loshchilov2017DecoupledWD} optimizer with an initial learning rate of 6e-5, 
% $\beta$ = 0.9, $\beta$ = 0.999, 
\jy{$\beta_1$ = 0.9, $\beta_2$ = 0.999} and a weight decay of 0.01 with four RTX 3090 GPUs. We employ a polynomial learning rate scheduling strategy with a power of 0.9. The batch size is set to 8, and the network is trained for 40 epochs.
% , including a 6-epoch warm-up phase. 
The VMamba backbone is initialized with pre-trained weights from ImageNet to enhance convergence and performance. Following previous methods, we employ the binary cross-entropy loss (BCE) to supervise the output maps of each level.

\begin{table*}[h]
  \centering
  \begin{tabular}{ccccccccc}
    \toprule
    \multirow{2}{*}{Methods} & \multicolumn{4}{c}{VMD-D Dataset} & \multicolumn{4}{c}{MMD Dataset} \\
    &IoU$\uparrow$&F$\beta\uparrow$&MAE$\downarrow$&Accuracy$\uparrow$
    &IoU$\uparrow$&F$\beta\uparrow$&MAE$\downarrow$&Accuracy$\uparrow$\\
    \midrule
    MINet & 0.412 & 0.676 & 0.148 & 0.854 & 0.623 & 0.810 & 0.169 & 0.833 \\
    PCSA & 0.193 & 0.464 & 0.198 & 0.803 & 0.619 & 0.806 & 0.175 & 0.824 \\
    PSPNet & 0.464 & 0.665 & 0.152 & 0.850 & 0.634 & 0.809 & 0.171 & 0.831 \\
    HFAN & 0.459 & 0.706 & 0.124 & 0.876 & 0.657 & 0.814 & 0.161 & 0.841 \\
    \midrule
    MirrorNet & 0.505 & 0.681 & 0.145 & 0.855 & 0.666 & 0.839 & 0.165 & 0.835 \\
    PMD-Net & 0.532 & 0.749 & 0.128 & 0.872 & 0.424 & 0.847 & 0.259 & 0.741 \\
    VCNet & 0.539 & 0.749 & 0.123 & 0.877 & - & - & - & -\\
    HetNet & 0.567 & 0.751 & 0.120 & 0.879 & 0.567 & 0.769 & 0.190 & 0.810 \\
    CSFWinformer & 0.557 & 0.759 & 0.113 & - & 0.718 & 0.809 & 0.148 & 0.852 \\
    PD-Net & 0.537 & 0.751 & 0.119 & 0.882 & 0.674 & 0.851 & 0.150 & 0.849 \\
    VMD-Net & 0.567 & \underline{0.787} & \underline{0.105} & \underline{0.895} & 0.723 & 0.812 & 0.146 & 0.854 \\
    MG-VMD & \underline{0.585} & 0.779 & 0.112 & 0.888 & \underline{0.725} & \underline{0.867} & \underline{0.127} & \underline{0.873} \\
    \midrule
    Ours & \textbf{0.646} & \textbf{0.804} & \textbf{0.095} & \textbf{0.905} & \textbf{0.793} & \textbf{0.872} & \textbf{0.095} & \textbf{0.905}\\
  \bottomrule
\end{tabular}
\caption{Quantitative comparison between MirrorMamba and state-of-the-art methods from relevant fields in video mirror detection. The best and the second results are shown in bold and underlined.}
\label{tab:vid}
\end{table*}

\begin{table}[h]
  \centering
  \begin{tabular}{ccccccc}
    \toprule
    \multirow{2}{*}{Methods} & \multicolumn{3}{c}{PMD Dataset} \\
    &IoU$\uparrow$&F$\beta\uparrow$&MAE$\downarrow$\\
    \midrule
    MINet & 0.604 & 0.760 & 0.037 \\
    VST & 0.591 & 0.769 & 0.035 \\
    UperNet & 0.685 & 0.838 & 0.025 \\
    \midrule
    MirrorNet & 0.585 & 0.741 & 0.043 \\
    PMD-Net & 0.660 & 0.794 & 0.032 \\
    VCNet & 0.640 & 0.815 & 0.032\\
    HetNet & 0.690 & 0.814 & 0.029 \\
    SATNet & 0.694 & \underline{0.847} & 0.025 \\
    CSFWinformer & \underline{0.700} & 0.838 & \underline{0.024} \\
    PD-Net & 0.653 & 0.793 & 0.031 \\
    \midrule
    Ours & \textbf{0.703} & \textbf{0.848} & \textbf{0.023}\\
  \bottomrule
\end{tabular}
\caption{Quantitative comparison between MirrorMamba and state-of-the-art methods from relevant fields in image-based mirror detection. The best and the second results are shown in bold and underlined. }
\label{tab:img}
\end{table}

\subsection{Datasets and Evaluation Metrics}
We evaluated our model on \sr{all video} mirror detection benchmarks, \sr{VMD-D and MMD.} VMD-D is a more challenging data set, consisting of 143 training videos (7,835 images) and 126 test videos (7,152 images). The dataset features complex scenes with numerous small mirrors, making detection particularly difficult. In contrast, the MMD dataset contains 18 training videos (4,653 images) and 19 test videos (5,074 images). Although the colors and mirrors in MMD are more uniform, the scenes are relatively monotonous, leading to a generally higher performance across models.

\sr{To show the robustness and \jy{extensibility} of \jy{MirrorMamba}, we also evaluated it on the most challenging and representative image-based mirror detection dataset, PMD. PMD consists of 5,096 training images and 571 test images, with great diversity in scenes and mirror appearances, making it more representative of real-world scenarios.}

\sr{In line with VMD-Net~\cite{Lin_2023_CVPR} and MG-VMD~\cite{Warren_2024_CVPR}, we preprocess the input images by cropping them to a size of 416$\times$416 and 224$\times$224 for VMD-D and MMD datasets respectively. For the PMD data set, we resize the input images to 512$\times$ 512, following the pre-processing steps adopted by previous works~\cite{10462920, 10.1609/aaai.v37i1.25157, Huang_Dong_Lin_Liu_W.H.Lau_Zuo_2023}. This ensures a fair comparison with state-of-the-art methods while maintaining consistency with their experimental setups.}

Following established practices \cite{Lin_2023_CVPR,Warren_2024_CVPR}, we evaluate our model using four metrics: Intersection over Union (IoU$\uparrow$) for quantifying spatial overlap between predicted and ground-truth segments, F-measure (F$\beta\uparrow$, $\beta^2$ is set to 0.3) for balancing precision and recall in binary classification, Mean Absolute Error (MAE$\downarrow$) for assessing pixel-wise error magnitude, and Accuracy$\uparrow$ for measuring overall prediction correctness. 
% F-measure (F$\beta\uparrow$) is calculated by:
% \begin{equation}\label{}F_\beta = \frac{(1 + \beta^2) \cdot \text{precision} \cdot \text{recall}}{\beta^2 \cdot \text{precision} + \text{recall}},\end{equation}
% where $\beta^2$ is set to 0.3 as suggested in previous work\cite{5206596}.

\begin{figure*}[h]
  \centering
  \includegraphics[width=\linewidth]{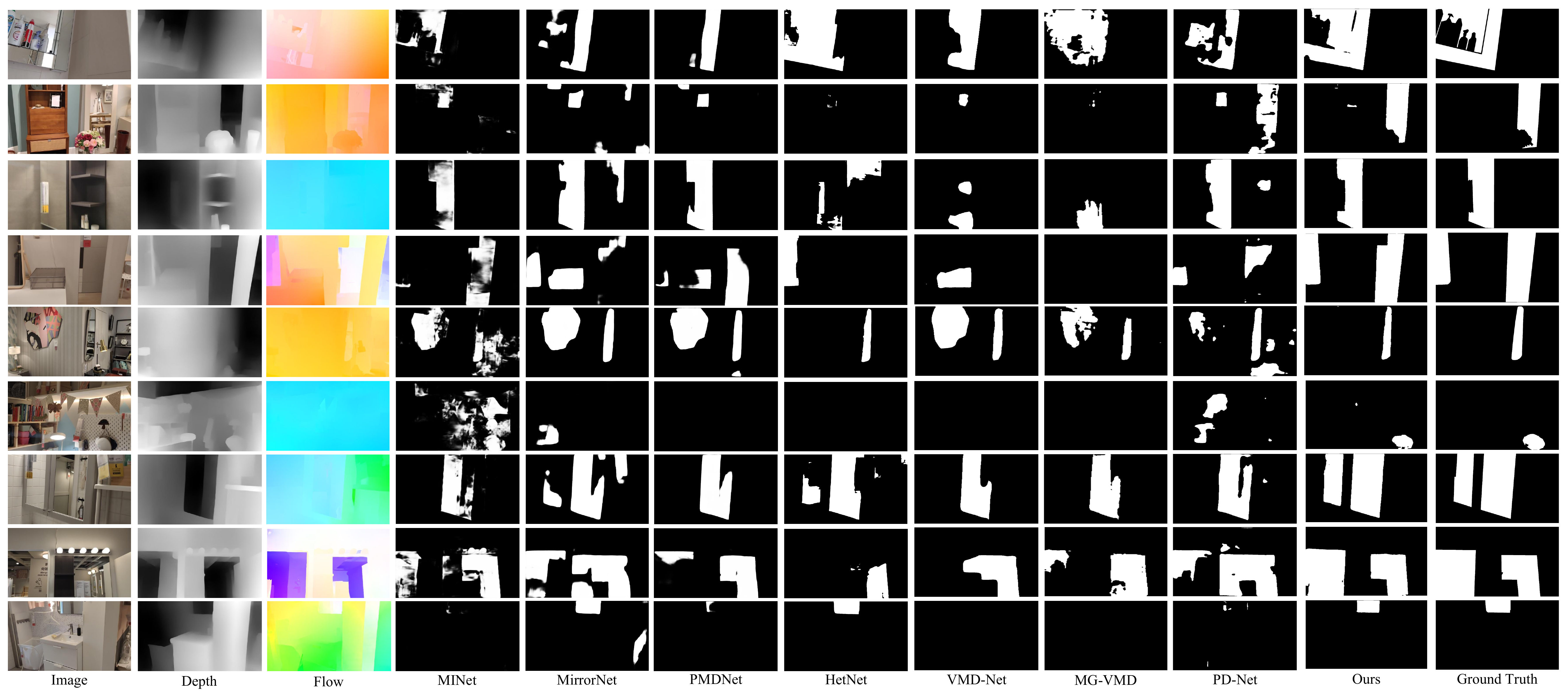}
  \caption{Qualitative results.
 %  table comparing our proposed method against MINet from salient object detection, 
 % MirrorNet, PMD-Net, HetNet from mirror detection, VMD-Net and MG-VMD from video mirror detection and PD-Net from RGB-D mirror detection.
 }
  \label{visual}
\end{figure*}

\subsection{Comparison on VMD-D and MMD}
We tested our models on the only two video mirror detection datasets. We compared our model with MINet~\cite{Pang_2020_CVPR} for salient object detection, PCSA~\cite{Gu_Wang_Wang_Liu_Cheng_Lu_2020} for video salient object detection; PSPNet~\cite{Zhao_2017_CVPR} for semantic segmentation;  HFAN~\cite{10.1007/978-3-031-19830-4_34} for video object segmentation; MirrorNet~\cite{Yang_2019_ICCV}, PMD-Net~\cite{Lin_2020_CVPR}, VCNet~\cite{9793716}, HetNet~\cite{10.1609/aaai.v37i1.25157}, SATNet~\cite{Huang_Dong_Lin_Liu_W.H.Lau_Zuo_2023} and CSFWinformer~\cite{10462920} for image mirror detection; PD-Net~\cite{Mei_2021_CVPR} for RGB-D mirror detection; and VMD-Net~\cite{Lin_2023_CVPR} and MG-VMD~\cite{Warren_2024_CVPR} for video mirror detection. Quantitative results, as shown in Table \ref{tab:vid}, demonstrate that our approach significantly outperforms all existing methods. 

\sr{We attribute the superior performance of MirrorMamba to its pioneering ability to fully leverage static cues in mirror detection, a key aspect overlooked by previous work. Previous video mirror detection methods beat image-based methods in video datasets by exploiting dynamic cues, but they ignore the use of static cues, limiting their ability to achieve better results.} The results in the video data sets fully demonstrate the superiority of our use of multiple cues.

% Furthermore, we compared our method with PD-Net~\cite{Mei_2021_CVPR}, the only RGB-D image mirror detection method, on MSD, PMD, VMD-D, and MMD datasets. We trained PD-Net using the relative depth maps generated by MiDaS alongside RGB images as input. The results show that our network achieves significantly better performance than PD-Net under the same conditions, further demonstrating its superiority. 

\subsection{Comparison on PMD}
We also compare our approach with state-of-the-art methods in the most representative image-based mirror detection dataset. Specifically, we remove the flow input and evaluate against MINet~\cite{Pang_2020_CVPR} and VST~\cite{Liu_2021_ICCV} for saliency object detection, UperNet~\cite{Xiao_2018_ECCV} for the semantic segmentation and MirrorNet~\cite{Yang_2019_ICCV}, PMD-Net~\cite{Lin_2020_CVPR}, VCNet~\cite{9793716}, HetNet~\cite{10.1609/aaai.v37i1.25157}, SATNet~\cite{Huang_Dong_Lin_Liu_W.H.Lau_Zuo_2023} and CSFWinformer~\cite{10462920} for mirror detection, PD-Net~\cite{Mei_2021_CVPR} for RGB-D mirror detection. Table \ref{tab:img} shows that our approach significantly outperforms all existing methods.

In the PMD dataset, we achieved the best results in all indicators and results comparable to the best model in the \jy{PMD} dataset. We attribute the superior performance of the MirrorMamba network to the use of multiple cues and the powerful modeling ability of Mamba. As the most challenging mirror detection dataset, we use multiple cues to cover all the conditions, making it the best result.

% \subsection{Comparison on RGBD-Mirror with PD-Net}
% Although our network is designed for relative depth, we also evaluated our method on the RGBD-Mirror dataset proposed by PD-Net, where our network also outperforms PD-Net by a substantial margin.

% \begin{table}[H]
% \centering
%   \begin{tabular}{lccc}
%     \toprule
%     Methods & IoU $\uparrow$ & F$\beta$ $\uparrow$ & MAE $\downarrow$ \\
%     \midrule
%     PD-Net & \textbf{0.778} & 0.825 & 0.042 \\
%     Ours & 0.777 & \textbf{0.886} & \textbf{0.035} \\
%     \bottomrule
%   \end{tabular}
%   \caption{Quantitative comparison between MirrorNet and PD-Net in \jy{RGBD}-Mirror.}
%   \label{tab:freq}
% \end{table}

% \noindent\textbf{Efficiency Analysis.}
% \sr{Our network, constructed by Mamba, achieves outstanding performance while maintaining high computational efficiency, as demonstrated in the table \ref{tab:efficiency}. For a consistent comparison, all the video mirror detection methods adopt an input resolution of 416$\times$416. Since MG-VMD integrates the optical flow computation core from RAFT~\cite{10.1007/978-3-030-58536-5_24} as part of its architecture, we exclude this module when calculating FLOPs and parameter counts to ensure fair evaluation.}

% \begin{table}[h]
% \centering
%   \begin{tabular}{cccc}
%     \toprule
%     Method & GFLOPs $\downarrow$ & Params.(M) $\downarrow$ \\
%     \midrule
%     VMD-Net & 165.82 & 62.2 \\
%     MG-VMD & 509.18 & 59.2 \\
%     Ours & 85.21 & 87.9 \\
%     \bottomrule
%   \end{tabular}
%   \caption{Efficiency comparison of VMD methods.}
%   \label{tab:efficiency}
% \end{table}

\subsection{Ablation Study} 
\sr{We conduct all the ablation experiments on VMD-D dataset as the most challenging video mirror detection dataset. In order to verify the effectiveness of multiple cues, we explored the impact of different cues on the results. Specifically, we arrange four groups of experiments: a). Correspondence: only RGB is used as input; b). Correspondence + relative depth: RGB + relative depth map is used as input; c). Correspondence + flow: RGB + optical flow map is used as input; d). Correspondence + relative depth + flow: RGB + relative depth + optical flow map is used as input. The result shown in Table \ref{tab:ablation1} demonstrates the effectiveness of our use of multiple cues. In addition, for the case of using only depth (a) and optical flow maps (b), our method also outperforms the corresponding competitors PD-Net~\cite{Mei_2021_CVPR} and MG-VMD~\cite{Warren_2024_CVPR}, respectively. We attribute this to the effectiveness of our module and the powerful modeling ability of Mamba. This further demonstrates the effectiveness of our method.}

\begin{table}[h]
\centering
  \begin{tabular}{ccccc}
    \toprule
     & IoU $\uparrow$ & F$\beta$ $\uparrow$ & MAE $\downarrow$  \\
    \midrule
    a) & 0.550 & 0.756 & 0.113 \\
    b) & 0.599 & 0.779 & 0.109 \\
    c) & 0.614 & 0.778 & 0.098 \\
    d) & \textbf{0.646} & \textbf{0.804} & \textbf{0.095} \\
    \bottomrule
  \end{tabular}
  \caption{Ablation study on the effectiveness of multiple cues.}
  \label{tab:ablation1}
\end{table}

To validate the effectiveness of our proposed modules, we conducted an ablation study on the Mamba-based Multi-direction Correspondence Extractor (MMCE) and the Layer-wise Boundary Enforcement Decoder (BED). We evaluated four configurations: (1) without MMCE and BED, (2) without MMCE but with BED, (3) with MMCE but without BED, and (4) with both MMCE and BED. The results, as shown in Table \ref{tab:ablation2}, demonstrate that the combination of MMCE and BED achieves the best performance, highlighting the complementary roles of these modules. MMCE effectively captures the correspondence and reflection properties of mirrors, while BED refines boundary details by leveraging high-level semantic information. This ablation study confirms that both modules are essential for robust mirror detection.

% \jyq{JY: the ablation study is too brief. You can discuss more details on each module, e.g. removing some of scanning angles in MMCE or replacing selective scan with plain concatenation. }

\begin{table}[h]
\centering
  \begin{tabular}{cccccc}
    \toprule
    \sr{MMCE} & BED & IoU $\uparrow$ & F$\beta$ $\uparrow$ & MAE $\downarrow$  \\
    \midrule
    $\times$ & $\times$ & 0.563 & 0.776 & 0.107 \\
    $\times$ & $\checkmark$ & 0.613 & 0.796 & 0.102 \\
    $\checkmark$ & $\times$ & 0.602 & 0.791 & 0.103 \\
    $\checkmark$ & $\checkmark$ & \textbf{0.646} & \textbf{0.804} & \textbf{0.095} \\
    \bottomrule
  \end{tabular}
  \caption{Ablation study on the effectiveness of \sr{MMCE} and BED on VMD-D.}
  \label{tab:ablation2}
\end{table}

\sr{Furthermore, we set up an experiment to evaluate the impact of different scanning methods on the results. We tried a) to set both scanning stages to horizontal scanning and b) to set both scanning stages to vertical scanning. The results in Table \ref{tab:ablation3} show that the use of scanning methods in different directions can improve the performance of the model.}

\begin{table}[h]
\centering
  \begin{tabular}{ccccc}
    \toprule
    scanning method & IoU $\uparrow$ & F$\beta$ $\uparrow$ & MAE $\downarrow$  \\
    \midrule
    horizontal & 0.635 & 0.797 & 0.104 \\
    vertical & 0.632 & 0.792 & 0.105 \\
    MirrorMamba & \textbf{0.646} & \textbf{0.804} & \textbf{0.095} \\
    \bottomrule
  \end{tabular}
  \caption{Ablation study on scanning strategy.}
  \label{tab:ablation3}
\end{table}

\section{Conclusion}
In this paper, we proposed a framework for mirror detection with multiple cues, including relative depth inconsistency, correspondence, and motion dynamics. We proposed MirrorMamba, the first unified framework designed for both image and video tasks. We designed the Mamba-based Multi-direction Correspondence Extractor (MMCE) to capture global symmetry relationships and the Mamba-based Layer-wise Boundary Enforcement Decoder (BED) to refine boundary details. Extensive experiments on both image and video mirror detection benchmarks demonstrate that our method achieves state-of-the-art performance. Additionally, we showcased the effectiveness of Mamba in mirror detection tasks, marking its first successful application in this field. 
% Our work provides a robust and efficient solution for mirror detection, with potential applications in augmented reality, autonomous driving, and scene understanding.
\bibliography{aaai2026}

@String(CVPR= {IEEE Conf. Comput. Vis. Pattern Recog.})

@String(ICCV= {Int. Conf. Comput. Vis.})

@String(ECCV= {Eur. Conf. Comput. Vis.})

@String(ICLR = {Int. Conf. Learn. Represent.})

@String(AAAI = {AAAI})

@String(CVPR  = {CVPR})

@String(ICCV  = {ICCV})

@String(ECCV  = {ECCV})

@String(ICLR  = {ICLR})

@InProceedings{Zhou_2017_CVPR,
author = {Zhou, Bolei and Zhao, Hang and Puig, Xavier and Fidler, Sanja and Barriuso, Adela and Torralba, Antonio},
title = {Scene Parsing Through ADE20K Dataset},
booktitle = {Proceedings of the IEEE Conference on Computer Vision and Pattern Recognition (CVPR)},
month = {July},
year = {2017}
}

@InProceedings{Costanzino_2023_ICCV,
    author    = {Costanzino, Alex and Ramirez, Pierluigi Zama and Poggi, Matteo and Tosi, Fabio and Mattoccia, Stefano and Di Stefano, Luigi},
    title     = {Learning Depth Estimation for Transparent and Mirror Surfaces},
    booktitle = {Proceedings of the IEEE/CVF International Conference on Computer Vision (ICCV)},
    month     = {October},
    year      = {2023},
    pages     = {9244-9255}
}

@InProceedings{Yang_2019_ICCV,
author = {Yang, Xin and Mei, Haiyang and Xu, Ke and Wei, Xiaopeng and Yin, Baocai and Lau, Rynson W.H.},
title = {Where Is My Mirror?},
booktitle = {Proceedings of the IEEE/CVF International Conference on Computer Vision (ICCV)},
month = {October},
year = {2019}
}

@InProceedings{Lin_2020_CVPR,
author = {Lin, Jiaying and Wang, Guodong and Lau, Rynson W.H.},
title = {Progressive Mirror Detection},
booktitle = {Proceedings of the IEEE/CVF Conference on Computer Vision and Pattern Recognition (CVPR)},
month = {June},
year = {2020}
}

@InProceedings{Guan_2022_CVPR,
    author    = {Guan, Huankang and Lin, Jiaying and Lau, Rynson W.H.},
    title     = {Learning Semantic Associations for Mirror Detection},
    booktitle = {Proceedings of the IEEE/CVF Conference on Computer Vision and Pattern Recognition (CVPR)},
    month     = {June},
    year      = {2022},
    pages     = {5941-5950}
}

@ARTICLE{9793716,
  author={Tan, Xin and Lin, Jiaying and Xu, Ke and Chen, Pan and Ma, Lizhuang and Lau, Rynson W.H.},
  journal={IEEE Transactions on Pattern Analysis and Machine Intelligence}, 
  title={Mirror Detection With the Visual Chirality Cue}, 
  year={2023},
  volume={45},
  number={3},
  pages={3492-3504},
  doi={10.1109/TPAMI.2022.3181030}}

@InProceedings{Lin_2023_CVPR,
    author    = {Lin, Jiaying and Tan, Xin and Lau, Rynson W.H.},
    title     = {Learning To Detect Mirrors From Videos via Dual Correspondences},
    booktitle = {Proceedings of the IEEE/CVF Conference on Computer Vision and Pattern Recognition (CVPR)},
    month     = {June},
    year      = {2023},
    pages     = {9109-9118}
}

@InProceedings{Warren_2024_CVPR,
    author    = {Warren, Alex and Xu, Ke and Lin, Jiaying and Tam, Gary K.L. and Lau, Rynson W.H.},
    title     = {Effective Video Mirror Detection with Inconsistent Motion Cues},
    booktitle = {Proceedings of the IEEE/CVF Conference on Computer Vision and Pattern Recognition (CVPR)},
    month     = {June},
    year      = {2024},
    pages     = {17244-17252}
}

@InProceedings{Mei_2021_CVPR,
    author    = {Mei, Haiyang and Dong, Bo and Dong, Wen and Peers, Pieter and Yang, Xin and Zhang, Qiang and Wei, Xiaopeng},
    title     = {Depth-Aware Mirror Segmentation},
    booktitle = {Proceedings of the IEEE/CVF Conference on Computer Vision and Pattern Recognition (CVPR)},
    month     = {June},
    year      = {2021},
    pages     = {3044-3053}
}

@misc{gu2024mambalineartimesequencemodeling,
      title={Mamba: Linear-Time Sequence Modeling with Selective State Spaces}, 
      author={Albert Gu and Tri Dao},
      year={2024},
      eprint={2312.00752},
      archivePrefix={arXiv},
      primaryClass={cs.LG},
      url={https://arxiv.org/abs/2312.00752}, 
}

@misc{wan2024sigmasiamesemambanetwork,
      title={Sigma: Siamese Mamba Network for Multi-Modal Semantic Segmentation}, 
      author={Zifu Wan and Pingping Zhang and Yuhao Wang and Silong Yong and Simon Stepputtis and Katia Sycara and Yaqi Xie},
      year={2024},
      eprint={2404.04256},
      archivePrefix={arXiv},
      primaryClass={cs.CV},
      url={https://arxiv.org/abs/2404.04256}, 
}

@ARTICLE{9178977,
  author={Ranftl, René and Lasinger, Katrin and Hafner, David and Schindler, Konrad and Koltun, Vladlen},
  journal={IEEE Transactions on Pattern Analysis and Machine Intelligence}, 
  title={Towards Robust Monocular Depth Estimation: Mixing Datasets for Zero-Shot Cross-Dataset Transfer}, 
  year={2022},
  volume={44},
  number={3},
  pages={1623-1637},
  doi={10.1109/TPAMI.2020.3019967}}

@InProceedings{Luo_2024_CVPR,
    author    = {Luo, Ao and Li, Xin and Yang, Fan and Liu, Jiangyu and Fan, Haoqiang and Liu, Shuaicheng},
    title     = {FlowDiffuser: Advancing Optical Flow Estimation with Diffusion Models},
    booktitle = {Proceedings of the IEEE/CVF Conference on Computer Vision and Pattern Recognition (CVPR)},
    month     = {June},
    year      = {2024},
    pages     = {19167-19176}
}

@article{Gu_Wang_Wang_Liu_Cheng_Lu_2020, title={Pyramid Constrained Self-Attention Network for Fast Video Salient Object Detection}, volume={34}, url={https://ojs.aaai.org/index.php/AAAI/article/view/6718}, DOI={10.1609/aaai.v34i07.6718}, abstractNote={&lt;p&gt;Spatiotemporal information is essential for video salient object detection (VSOD) due to the highly attractive object motion for human’s attention. Previous VSOD methods usually use Long Short-Term Memory (LSTM) or 3D ConvNet (C3D), which can only encode motion information through step-by-step propagation in the temporal domain. Recently, the non-local mechanism is proposed to capture long-range dependencies directly. However, it is not straightforward to apply the non-local mechanism into VSOD, because i) it fails to capture motion cues and tends to learn motion-independent global contexts; ii) its computation and memory costs are prohibitive for video dense prediction tasks such as VSOD. To address the above problems, we design a &lt;em&gt;Constrained Self-Attention&lt;/em&gt; (CSA) operation to capture motion cues, based on the prior that objects always move in a continuous trajectory. We group a set of CSA operations in &lt;em&gt;Pyramid&lt;/em&gt; structures (PCSA) to capture objects at various scales and speeds. Extensive experimental results demonstrate that our method outperforms previous state-of-the-art methods in both accuracy and speed (110 FPS on a single Titan Xp) on five challenge datasets. Our code is available at https://github.com/guyuchao/PyramidCSA.&lt;/p&gt;}, number={07}, journal={Proceedings of the AAAI Conference on Artificial Intelligence}, author={Gu, Yuchao and Wang, Lijuan and Wang, Ziqin and Liu, Yun and Cheng, Ming-Ming and Lu, Shao-Ping}, year={2020}, month={Apr.}, pages={10869-10876} }

@ARTICLE{10462920,
  author={Xie, Zhifeng and Wang, Sen and Yu, Qiucheng and Tan, Xin and Xie, Yuan},
  journal={IEEE Transactions on Image Processing}, 
  title={CSFwinformer: Cross-Space-Frequency Window Transformer for Mirror Detection}, 
  year={2024},
  volume={33},
  number={},
  pages={1853-1867},
  doi={10.1109/TIP.2024.3372468}}

@InProceedings{Hu_2018_CVPR,
author = {Hu, Jie and Shen, Li and Sun, Gang},
title = {Squeeze-and-Excitation Networks},
booktitle = {Proceedings of the IEEE Conference on Computer Vision and Pattern Recognition (CVPR)},
month = {June},
year = {2018}
}

@InProceedings{10.1007/978-3-031-19830-4_34,
author="Pei, Gensheng
and Shen, Fumin
and Yao, Yazhou
and Xie, Guo-Sen
and Tang, Zhenmin
and Tang, Jinhui",
editor="Avidan, Shai
and Brostow, Gabriel
and Ciss{\'e}, Moustapha
and Farinella, Giovanni Maria
and Hassner, Tal",
title="Hierarchical Feature Alignment Network for Unsupervised Video Object Segmentation",
booktitle="Computer Vision -- ECCV 2022",
year="2022",
publisher="Springer Nature Switzerland",
address="Cham",
pages="596--613",
isbn="978-3-031-19830-4"
}

@InProceedings{Pang_2020_CVPR,
author = {Pang, Youwei and Zhao, Xiaoqi and Zhang, Lihe and Lu, Huchuan},
title = {Multi-Scale Interactive Network for Salient Object Detection},
booktitle = {Proceedings of the IEEE/CVF Conference on Computer Vision and Pattern Recognition (CVPR)},
month = {June},
year = {2020}
}

@InProceedings{Liu_2021_ICCV,
    author    = {Liu, Nian and Zhang, Ni and Wan, Kaiyuan and Shao, Ling and Han, Junwei},
    title     = {Visual Saliency Transformer},
    booktitle = {Proceedings of the IEEE/CVF International Conference on Computer Vision (ICCV)},
    month     = {October},
    year      = {2021},
    pages     = {4722-4732}
}

@InProceedings{Xiao_2018_ECCV,
author = {Xiao, Tete and Liu, Yingcheng and Zhou, Bolei and Jiang, Yuning and Sun, Jian},
title = {Unified Perceptual Parsing for Scene Understanding},
booktitle = {Proceedings of the European Conference on Computer Vision (ECCV)},
month = {September},
year = {2018}
}

@InProceedings{Zhao_2017_CVPR,
author = {Zhao, Hengshuang and Shi, Jianping and Qi, Xiaojuan and Wang, Xiaogang and Jia, Jiaya},
title = {Pyramid Scene Parsing Network},
booktitle = {Proceedings of the IEEE Conference on Computer Vision and Pattern Recognition (CVPR)},
month = {July},
year = {2017}
}

@inproceedings{10.1609/aaai.v37i1.25157,
author = {He, Ruozhen and Lin, Jiaying and Lau, Rynson W.H.},
title = {Efficient mirror detection via multi-level heterogeneous learning},
year = {2023},
isbn = {978-1-57735-880-0},
publisher = {AAAI Press},
url = {https://doi.org/10.1609/aaai.v37i1.25157},
doi = {10.1609/aaai.v37i1.25157},
booktitle = {Proceedings of the Thirty-Seventh AAAI Conference on Artificial Intelligence and Thirty-Fifth Conference on Innovative Applications of Artificial Intelligence and Thirteenth Symposium on Educational Advances in Artificial Intelligence},
articleno = {88},
numpages = {9},
series = {AAAI'23/IAAI'23/EAAI'23}
}

@inproceedings{a7242453d69a4c92b560a4acd7ba9bf7,  title     = "ZOOM: Learning Video Mirror Detection with Extremely-Weak Supervision", author    = "Ke Xu and Siu, {Tsun Wai} and Lau, {Rynson W.H.}",  year      = "2024",  doi       = "10.1609/aaai.v38i6.28450",  language  = "English",  isbn      = "1-57735-887-2",  series    = "Proceedings of the AAAI Conference on Artificial Intelligence",  publisher = "AAAI Press",  number    = "6",  pages     = "6315--6323",  editor    = "Michael Wooldridge and Jennifer Dy and Sriraam Natarajan",  booktitle = "Proceedings of the 38th AAAI Conference on Artificial Intelligence",  note      = "38th AAAI Conference on Artificial Intelligence, AAAI 2024 ; Conference date: 20-02-2024 Through 27-02-2024", }

@article{dosovitskiy2020vit,
  title={An Image is Worth 16x16 Words: Transformers for Image Recognition at Scale},
  author={Dosovitskiy, Alexey and Beyer, Lucas and Kolesnikov, Alexander and Weissenborn, Dirk and Zhai, Xiaohua and Unterthiner, Thomas and  Dehghani, Mostafa and Minderer, Matthias and Heigold, Georg and Gelly, Sylvain and Uszkoreit, Jakob and Houlsby, Neil},
  journal={ICLR},
  year={2021}
}

@inproceedings{liu2021Swin,
  title={Swin Transformer: Hierarchical Vision Transformer using Shifted Windows},
  author={Liu, Ze and Lin, Yutong and Cao, Yue and Hu, Han and Wei, Yixuan and Zhang, Zheng and Lin, Stephen and Guo, Baining},
  booktitle={Proceedings of the IEEE/CVF International Conference on Computer Vision (ICCV)},
  year={2021}
}

@article{liu2024vmamba,
  title={VMamba: Visual State Space Model},
  author={Liu, Yue and Tian, Yunjie and Zhao, Yuzhong and Yu, Hongtian and Xie, Lingxi and Wang, Yaowei and Ye, Qixiang and Liu, Yunfan},
  journal={arXiv preprint arXiv:2401.10166},
  year={2024}
}

@article{Huang_Dong_Lin_Liu_W.H.Lau_Zuo_2023,
title={Symmetry-Aware Transformer-Based Mirror Detection}, 
volume={37}, 
url={https://ojs.aaai.org/index.php/AAAI/article/view/25173}, 
DOI={10.1609/aaai.v37i1.25173}, 
abstractNote={Mirror detection aims to identify the mirror regions in the given input image. Existing works mainly focus on integrating the semantic features and structural features to mine specific relations between mirror and non-mirror regions, or introducing mirror properties like depth or chirality to help analyze the existence of mirrors. In this work, we observe that a real object typically forms a loose symmetry relationship with its corresponding reflection in the mirror, which is beneficial in distinguishing mirrors from real objects. Based on this observation, we propose a dual-path Symmetry-Aware Transformer-based mirror detection Network (SATNet), which includes two novel modules: Symmetry-Aware Attention Module (SAAM) and Contrast and Fusion Decoder Module (CFDM). Specifically, we first adopt a transformer backbone to model global information aggregation in images, extracting multi-scale features in two paths. We then feed the high-level dual-path features to SAAMs to capture the symmetry relations. Finally, we fuse the dual-path features and refine our prediction maps progressively with CFDMs to obtain the final mirror mask. Experimental results show that SATNet outperforms both RGB and RGB-D mirror detection methods on all available mirror detection datasets.}, 
number={1}, 
journal={Proceedings of the AAAI Conference on Artificial Intelligence}, 
author={Huang, Tianyu and Dong, Bowen and Lin, Jiaying and Liu, Xiaohui and W.H. Lau, Rynson and Zuo, Wangmeng}, 
year={2023}, 
month={Jun.}, 
pages={935-943} }

@inproceedings{Loshchilov2017DecoupledWD,
  title={Decoupled Weight Decay Regularization},
  author={Ilya Loshchilov and Frank Hutter},
  booktitle={International Conference on Learning Representations},
  year={2017},
  url={https://api.semanticscholar.org/CorpusID:53592270}
}

@article{paszke2019pytorch,
  title={Pytorch: An imperative style, high-performance deep learning library},
  author={Paszke, Adam and Gross, Sam and Massa, Francisco and Lerer, Adam and Bradbury, James and Chanan, Gregory and Killeen, Trevor and Lin, Zeming and Gimelshein, Natalia and Antiga, Luca and others},
  journal={Advances in neural information processing systems},
  volume={32},
  year={2019}
}

@InProceedings{Lin_2023_ICCV,
    author    = {Lin, Jiaying and Lau, Rynson W.H.},
    title     = {Self-supervised Pre-training for Mirror Detection},
    booktitle = {Proceedings of the IEEE/CVF International Conference on Computer Vision (ICCV)},
    month     = {October},
    year      = {2023},
    pages     = {12227-12236}
}

@ARTICLE{10595128,
  author={Zha, Mingfeng and Fu, Feiyang and Pei, Yunqiang and Wang, Guoqing and Li, Tianyu and Tang, Xiongxin and Yang, Yang and Tao Shen, Heng},
  journal={IEEE Transactions on Circuits and Systems for Video Technology}, 
  title={Dual Domain Perception and Progressive Refinement for Mirror Detection}, 
  year={2024},
  volume={34},
  number={11},
  pages={11942-11953},
  doi={10.1109/TCSVT.2024.3426673}}

\end{document}